\documentclass[runningheads]{llncs}

\usepackage[colorlinks=true]{hyperref}
\usepackage{amsmath}
\usepackage{graphicx}
\usepackage{booktabs}
\usepackage{amsfonts}
\usepackage{amssymb}
\usepackage[ruled,vlined]{algorithm2e}
\usepackage{caption}
\usepackage{subcaption}
\usepackage{xcolor}

\newcommand{\bm}{\boldmath}

\newcommand{\xx}{\mathbf{x}}
\newcommand{\yy}{\mathbf{y}}

\newcommand{\sss}{\mathbf{s}}
\newcommand{\real}{\mathbb{R}}
\newcommand{\appen}{\textcolor{red}{Appendix}\xspace}
\newcommand{\ourmethod}{CRaC\xspace}
\usepackage{xcolor}
\definecolor{brightgray}{RGB}{220,220,220}

\newcommand{\N}{\mathbb{N}}

\def\vrho{{\boldsymbol{\rho}}}
\def\vlambda{{\boldsymbol{\lambda}}}
\def\rvl{{\mathbf{l}}}

\begin{document}
\title{Class and Region-Adaptive Constraints\\ for Network Calibration}
%
%

\author{Balamurali Murugesan \and
Julio Silva-Rodriguez \and
\\ Ismail Ben Ayed \and
Jose Dolz}
\authorrunning{Murugesan \textit{et al.}}
%
\institute{ETS Montreal}
\maketitle              

\begin{abstract}
  
In this work, we present a novel approach to calibrate segmentation networks that considers the inherent challenges posed by different categories and object regions. In particular, we present a formulation that integrates class and region-wise constraints into the learning objective, with multiple penalty weights to account for class and region differences. Finding the optimal penalty weights manually, however, might be unfeasible, and potentially hinder the optimization process. To overcome this limitation, we propose an approach based on Class and Region-Adaptive constraints (CRaC), which allows to learn the class and region-wise penalty weights during training. CRaC is based on a general Augmented Lagrangian method, a well-established technique in constrained optimization. Experimental results on two popular segmentation benchmarks, and two well-known segmentation networks, demonstrate the superiority of CRaC compared to existing approaches. The code is available at: \url{https://github.com/Bala93/CRac/}

\end{abstract}

\section{Introduction}

Despite the remarkable progress achieved by deep neural networks (DNNs), they are susceptible to suffer from miscalibration, leading to 
overconfident 
predictions 
\cite{guo2017calibration,minderer2021revisiting}, even when they are incorrect. This issue becomes especially significant in safety-critical scenarios, such as medical diagnosis or treatment, where producing accurate uncertainty estimates is of paramount importance. An inherent cause of network miscalibration is known to be the implicit bias for low-entropy predictions caused by popular supervised losses, such as the cross-entropy, which encourages large differences between the logit of the ground truth category and the remaining classes \cite{mukhoti2020calibrating}.

A myriad of approaches have emerged to mitigate network miscalibration, which mainly focus on either post-processing strategies or integrating additional learning objectives during training. The first family of approaches, i.e., \textit{post-processing} methods, offers a simple alternative for modifying the softmax predictions in a post-hoc fashion by establishing a mapping from raw network outputs to well-calibrated confidences
\cite{Ding2021LocalTemp,guo2017calibration,gupta2020calibration,Tomani2021Posthoc,zhang2020mix}. The second category involves incorporating additional regularization during training, typically penalizing low-entropy predictions. For example, \cite{pereyra2017regularizing} introduced an explicit term that maximizes the Shannon entropy of the network predictions during training, which was later extended in \cite{larrazabal2023maximum} by penalizing low-entropy distributions only in incorrect predictions. Furthermore, popular losses for classification, such as Label smoothing \cite{szegedy2016rethinking} or focal loss \cite{lin2017focal}, implicitly integrate an entropy maximization term, which has a favourable effect on calibration \cite{mukhoti2020calibrating,muller2019does}. More recently, \cite{liu2022devil,liu2023class,murugesan2022calibrating} propose to enforce inequality constraints on the logit space, allowing to control the margin on logit distances, ultimately reducing overconfidence in the predictions. This provided more flexibility than systematically maximizing the entropy of the predictions, as in \cite{mukhoti2020calibrating,muller2019does}, which results in gradients that continually push towards a non-informative solutions. Other works include the integration of pair-wise constraints between classes \cite{cheng2022calibrating} or augmenting the training dataset by convex combinations of random pairs of images and their associated labels, e.g., MixUp \cite{thulasidasan2019mixup}. Nevertheless, even though these works have achieved remarkable progress in addressing miscalibration in both classification \cite{cheng2022calibrating,guo2017calibration,gupta2020calibration,mukhoti2020calibrating,muller2019does,pereyra2017regularizing,Tomani2021Posthoc} and segmentation tasks \cite{Ding2021LocalTemp,larrazabal2023maximum,liu2022devil,murugesan2022calibrating}, they disregard neighbour pixel relationships, in terms of classes, which is of significant relevance in semantic image segmentation.   
 
Certainly, one of the factors contributing to the reduced performance of these losses in segmentation tasks arises from the uniform, or near-to-uniform, distribution enforced in the network predictions (whether logit or softmax predictions), which neglects the spatial context \cite{murugesan2023trust}. To overcome this issue, and to integrate class-wise information of the surrounding pixels during training, Spatially Varying Label Smoothing (SVLS) \cite{islam2021spatially} introduced a label smoothing strategy that captures the structural uncertainty required in semantic segmentation. More specifically, SVLS uses a Gaussian kernel applied across the one-hot encoded ground truth, leading to class probabilities based on a soft combination of neighboring pixels. As exposed in \cite{murugesan2023trust,murugesan2024neighbor}, SVLS integrates an implicit penalty on softmax predictions, which enforces a prior based on soft class proportions of surrounding pixels. This strategy, however, lacks a mechanism to control the influence of the constraint over the main objective, potentially hindering the
optimization process. To circumvent this limitation, authors presented a simple solution that combines the standard cross-entropy with an explicit penalty, where both the prior and its impact can be easily controlled. 

Although the work proposed in \cite{murugesan2023trust,murugesan2024neighbor}  achieves greater calibration performance than existing alternatives, and integrates class-relationships across a pixel and its neighbours, it presents two major limitations: \textit{1)} The scalar balancing weight that controls the importance of the penalty is equal for all classes, and for all the regions. This scenario is suboptimal, as it can hamper the network performance when some classes are more challenging to segment, or under-represented. 
Furthermore, this strategy considers than the weight of the penalty should be the same for a pixel inside the object (likely to have \textit{low uncertainty}) than for a pixel within the organ boundaries (likely to have \textit{high uncertainty}). \textit{2)} The value of the balancing weight is defined before network optimization, lacking an adaptive strategy during training. For example, as the training evolves, the cross-entropy loss pushes towards lower-entropy predictions, whereas the penalty weight is the same at the beginning and the end of the training. 

Based on these findings, we can summarize our contributions as:

\begin{enumerate}
    \item We propose a class and region-wise constraint approach to tackle the miscalibration issue in semantic segmentation models. In particular, we formulate a solution that considers the specificities of each category and different regions by introducing independent class and region-wise penalty weights. This contrasts with the prior work in \cite{murugesan2023trust}, where a uniform scalar penalty weight is employed, regardless of categories or regions.
    \item Furthermore, we transfer the constrained problem to its dual unconstrained optimization counterpart by using an Augmented Lagrangian method (ALM). This alleviates the need for manually adjusting each penalty weight and allows, through a series of iterative \textit{inner} and \textit{outer} steps, to find the optimal value of each penalty weight, which can be learned in an adaptive manner.  
    \item Comprehensive experiments on two popular segmentation benchmarks, and with two well-known segmentation backbones, demonstrate the superiority of our approach over a set of relevant recent calibration approaches. 
\end{enumerate}

\section{Methodology}

\noindent \textbf{Notation}. We denote the training dataset as $\mathcal{D}(\mathcal{X}, \mathcal{Y})=\{(\xx^{(n)}, \yy^{(n)})\}_{n=1}^N$, where $\xx^{(n)} \in \mathcal{X} \subset \mathbb{R}^{\Omega_n}$ represents the $n^{th}$ image, $\Omega_n$ its spatial image domain, and $\yy^{(n)} \in \mathcal{Y} \subset \mathbb{R}^K$ the corresponding pixel-wise ground-truth annotation with $K$ classes, which is provided as a one-hot encoding vector. Given an input image $\xx^{(n)}$, a neural network parameterized by $\boldsymbol{\theta}$ generates a logit vector $f_{\boldsymbol{\theta}}(\xx^{(n)})= \rvl^{(n)} \in \mathbb{R}^{\Omega_n \times K}$, which can be converted into probability values with the softmax operator, $\text{softmax}(\rvl^{(n)})=\mathbf{s}^{(n)} \in [ 0,1 ]^{\Omega_n \times K}$. 
To simplify the notations, we omit sample indices, as this does not lead to any ambiguity. 

\subsection{Background}
Despite its importance in dense prediction tasks, such as segmentation, very few approaches consider pixel spatial relationships across classes to address the miscalibration issue. Spatially Varying Label Smoothing (SVLS) \cite{islam2021spatially} integrates neighbour class information by softening the pixel label assignments with a discrete spatial Gaussian kernel. More recently, NACL \cite{murugesan2023trust,murugesan2024neighbor} formally showed that SVLS actually enforces an implicit constraint on soft class proportions of surrounding pixels, and propose the following constrained optimization problem to alleviate the limitations of SVLS: 
 
\begin{align}
\label{eq:constrained}
\min_{\boldsymbol{\theta}} \quad \mathcal{L}_{CE} \quad \textrm{s.t.} \quad  \bm{\tau} = \rvl, 
\end{align}

which can be approximated by incorporating an explicit penalty, whose overall learning objective 
is defined as: 

    
\begin{equation}
\label{eq:proposed-NACL}
\min_{\boldsymbol{\theta}} \quad \sum_{i \in \Omega}\sum_{k \in K}(-y_k^{(i)}\log (s_k^{(i)}) + \lambda  | \tau_k^{(i)} - l_k^{(i)}|).
\end{equation}

\noindent The first term in the above equation is the standard cross-entropy loss on a given pixel, the second term is a linear penalty over the pixel logit distributions, $\boldsymbol{\tau}$ is a 
prior, and $\lambda$ the balancing hyperparameter that controls the importance of each term. With this objective,
when the constraint $|\tau_k - l_k|$ deviates from $0$ (i.e., $\tau_k$ and $l_k$ are different) the value of the penalty term increases. Thus, as the prior $\mathbf{\boldsymbol{\tau}}=\{\tau_0,...,\tau_k\}$ captures the class distribution of a 2D patch\footnote{More details about the priors and the enforced constraint in \cite{murugesan2023trust,murugesan2024neighbor}.} surrounding the pixel, the penalty enforces the predicted logit distribution $\rvl$ to follow 
$\boldsymbol{\tau}$.

\subsection{Class and region-wise penalties}

The unconstrained formulation presented in Equation \ref{eq:proposed-NACL} employs a single uniform penalty. We argue that this scenario is suboptimal, as it disregards differences across individual categories, or even different regions with different uncertainty in the target object, which may pose distinct inherent learning challenges. For example, annotations from a patch in the center of an organ typically have less uncertainty that labels in within the organ boundaries. A better, and more optimal strategy would integrate multiple penalty weights $\lambda$, one for each category and type of patch/region, leading to a set of penalty weights $\mathbf{\Lambda} \in \mathbb{R}_{+}^{K \times R}$, with $R$ being the number of regions. For simplicity, in this work we will consider only two types of regions (i.e., $R=2$, leading to $\mathbf{\Lambda}=\{\vlambda_0,\vlambda_1\}$), that we denote as \textit{inner} and \textit{outer} regions, and whose sets are defined as $\mathcal{I}$ and $\mathcal{O}$, respectively. More concretely, if the surrounding ground truth patch of a given pixel only contains one category, it will be considered as an \textit{inner} patch, whereas otherwise it will be an \textit{outer} patch. Thus, we can formally define our 
formulation as:

\begin{align}
\label{eq:proposed-ours}
\min_\theta \quad \sum_{i \in \Omega} \mathcal{H}(\yy^{(i)},\sss^{(i)}) + \sum_{i \in \mathcal{I}}\sum_{k \in K}\mathbf{\lambda}_{k,0}  | \tau_k^{(i)} - l_k^{(i)}| + \sum_{i \in \mathcal{O}}\sum_{k \in K} \mathbf{\lambda}_{k,1}  | \tau_k^{(i)} - l_k^{(i)}|,
\end{align}

where $\mathcal{H}(\yy,\sss)$ is the standard cross-entropy loss. As stated in prior literature in constrained convolutional neural networks \cite{marquez2017imposing,rony2021augmented,liu2023class,silva2023closer}, while $\mathbf{\Lambda^*}\in \real^{K \times R}_{+}$ are the Lagrange multipliers of the presented problem, and $\mathbf{\Lambda} = \mathbf{\Lambda^*}$ could be considered the best choice to solve (\ref{eq:proposed-ours}), using $\mathbf{\Lambda^*}$ as the penalty weights may not feasible in practice. On the other hand, finding the optimal value for each penalty weight manually can pose optimization challenges, particularly for datasets with a large number of classes. 

\subsection{The proposed class and region adaptive solution}

\noindent \textbf{General Augmented Lagrangian.} To alleviate the need of having to chose the penalty weights $\mathbf{\Lambda}\in \real^{K \times R}_{+}$, we propose to use an Augmented Lagrangian Multiplier (ALM) method. ALM approaches are optimization techniques that integrate penalties and primal-dual updates to efficiently tackle constrained optimization problems. These methods iteratively refine solutions by adjusting penalty terms based on Lagrange multipliers, effectively balancing between satisfying constraints, i.e., the penalties, and minimizing the main objective function, in our case the cross-entropy loss. ALM approaches are favoured due to their ability to handle complex constraints and their robust performance across various optimization scenarios, and enjoy widespread popularity in the general context of optimization \cite{bertsekas1996constrained,nocedal2006numerical}. A general constrained optimization problem can be formally defined as:

\begin{equation}
\label{eq:general_ALM}
 \min_x \quad g(x) \quad \text{s.t.} \quad h_i(x)\leq 0, \quad i=1,\dots,n
\end{equation}

\noindent with $g:\real^d\rightarrow \real$ the \textit{objective function} and $h_i:\real^d\rightarrow\real, i=1,\dots,n$ being the \textit{set of constraint functions}. Generally, this problem is tackled by solving a succession of $j\in \mathbb{N}$ unconstrained problems, each solved approximately w.r.t $x$:

\begin{equation}
\label{eq:lagrangian_alm}
    \min_{x,\lambda} \quad \mathcal{L}^{(j)}(x) = g(x) + \sum_{i=1}^n P(h_i(x), \rho_i^{(j)}, \lambda_i^{(j)}),
\end{equation}

\noindent where $P:\real \times \real_{++}\times \real_{++} \rightarrow \real$ is a \textit{penalty-Lagrangian function}, whose derivative w.r.t. its first variable $P'(z,\rho,\lambda) \equiv \frac{\partial}{\partial z}P(z,\rho,\lambda)$ exists, is positive and continuous for all $z \in \real$ and $(\rho, \lambda) \in (\real_{++})^2$.  
In addition, we denote $\vrho^{(j)}=(\rho^{(j)}_i)_{1\leq i\leq n}\in \real_{++}^n$ and $\vlambda^{(j)}=(\lambda^{(j)}_i)_{1\leq i\leq n}\in\real_{++}^n$ as the penalty parameters and multipliers associated to the penalty $P$ at the iteration $\textit{j}$. We detail in the \appen the set of axioms that any penalty function $P$ must satisfy \cite{birgin2005numerical}. 

The ALM can be split into two iterations. First, in the \emph{outer} iterations, which indexed by $j$, the \textit{penalty multipliers} $\vlambda$ and the \textit{penalty parameters} $\vrho$ are updated. Then, during the \emph{inner} iterations, the objective $\mathcal{L}^{(j)}$ (Eq \ref{eq:lagrangian_alm}) is minimized using the previous solution as initialization to this problem. Particularly, the penalty multipliers $\vlambda^{(j)}$ are updated to the derivative of $P$ w.r.t. to the solution obtained during the last \emph{inner} step:

\begin{equation}
    \label{eq:lambda_update}
    \lambda_i^{(j+1)}=P'(h_i(x), \rho_i^{(j)}, \lambda_i^{(j)}).
\end{equation}

\noindent This approach increases the value of the penalty multipliers when the constraint is violated, and decreases their value otherwise. Thus, integrating an ALM during optimization enables an \emph{adaptive} and \emph{learnable} strategy to determine an optimal value for the penalty weights. 


\noindent \textbf{Our global learning objective.} Based on the benefits detailed above, we propose to solve the problem in Eq. \ref{eq:proposed-ours} by using an ALM approach. More concretely, we reformulate this problem by integrating a penalty function $P$, which is parameterized by $(\vrho, \vlambda)\in\real_{++}^K\times\real_{++}^K$: 

\begin{align}
\label{eq:final_problem}
\min_{\theta,\vlambda_0,\vlambda_1} \quad \sum_{i \in \Omega} \mathcal{H}(\yy^{(i)},\sss^{(i)}) &+ \sum_{i \in \mathcal{I}}\sum_{k \in K} P(\tau_k^{(i)} - l_k^{(i)},\rho_{k,0}, \lambda_{k,0})  \nonumber \\ &+ \sum_{i \in \mathcal{O}}\sum_{k \in K} P(\tau_k^{(i)} - l_k^{(i)},\rho_{k,1}, \lambda_{k,1}).
\end{align}

To obtain an accurate estimate of the penalty multipliers at each epoch, we compute the satisfaction of the constraint on the validation set, following standard practices in machine learning. In this work, we consider that a single training epoch approximately minimizes the loss function. Then, we compute the average penalty multiplier on the validation set. This means that, after a training epoch $j$, the penalty multipliers for all $k=1,...,K$ and each region $r$ at epoch $j+1$ can be computed as:

\begin{equation}
    \lambda_{k,r}^{(j+1)} = \frac{1}{|\mathcal{D}_{val}|}\sum_{(\xx,\yy)\in\mathcal{D}_{val}} P'\left(\tau_k-l_k, \rho_{k,r}^{(j)}, \lambda_{k,r}^{(j)}\right).
\end{equation}

Furthermore, $\rho$ is updated as:

\begin{equation}
    \rho_{k,r}^{(j+1)} =
    \begin{cases}
        \gamma \rho_{k,r}^{(j)} \qquad |\tau_k^{(i)} - l_k^{(i)}| > \mu * |\tau_k^{(i)} - l_k^{(i)}|;\\
        \rho_{k,r}^{(j)} \qquad \text{otherwise},
    \end{cases}
\end{equation}

\noindent where $\mu$ is a constant factor that determines the amount of the update. 
Last, following prior works on ALM in the context of constrained CNNs \cite{liu2023class,rony2021augmented,silva2023closer}, we employ PHR as the penalty, which is defined as:

\begin{equation}
\label{eq:phr}
    \mathrm{PHR}(z,\rho,\lambda) =
    \begin{cases}
        \lambda z + \frac12 \rho z^2 \quad &\text{if} \quad \lambda + \rho z \geq 0;\\
        -\frac{\lambda^2}{2\rho} \quad &\text{otherwise}.
    \end{cases}
\end{equation}

\section{Experiments}


\noindent \textbf{\textit{Datasets.}} Following NACL \cite{murugesan2023trust}, we use the ACDC and FLARE datasets with its setting. \textbf{
    ACDC} \cite{bernard2018deep} contains 100 patient exams with cardiac MR volumes and their respective 
    pixel-wise annotations. 
    We follow the standard practices on this dataset, and extract 2D slices from the 
    volumes, which are resized to 224$\times$224. Furthermore, \textbf{
    FLARE} \cite{AbdomenCT-1K} includes $360$ volumes of multiple organs in abdominal CTs, together with their corresponding segmentation masks, which are resampled to a common space and cropped to 192$\times$192$\times$30. 

\noindent \textbf{\textit{Baselines.}} 
We compare to 
relevant calibration losses, as well as to state-of-the-art methods for calibration in medical imaging segmentation: Focal Loss (FL) \cite{mukhoti2020calibrating}, penalizing low-entropies (ECP) \cite{pereyra2017regularizing}, Label smoothing (LS) \cite{szegedy2016rethinking}, SVLS \cite{islam2021spatially}, MbLS \cite{liu2022devil}, NACL \cite{murugesan2023trust} and BWCR \cite{karani2023boundary}. As segmentation backbones, we have selected two well-known and popular networks, UNet \cite{ronneberger2015u} and nnUNet \cite{nnUNet}.

\noindent \textbf{\textit{Implementation details.}} For most of the compared methods, we use the hyperparameters values reported in \cite{murugesan2023trust}:  FL ($\gamma=3$), LS ($\alpha=0.1$), ECP ($\lambda=0.1$), MbLS ($\lambda=0.1$ and $m=10$), SVLS ($\sigma=2$) and NACL ($\lambda=0.1$). Furthermore, for BWCR, the impact of the logit consistency is controlled by $\lambda_{min}=0.01$, and $\lambda_{max}=1$. Regarding the prior used in NACL and our method \ourmethod, we use the one proposed in \cite{murugesan2023trust}, which is defined as $\tau_k = \sum_{\substack{i=1}}^d y^k_i$, which is computed over a 3$\times$3 patch. We train all the models during 100 epochs, with ADAM \cite{kingma2014adam} as optimizer and a batch size fixed to 16. The learning rate is set to 10$^{-3}$ for the first 50 epochs, and reduced to 10$^{-4}$ afterwards. Following \cite{murugesan2023trust}, the models are trained on 2D slices, and the evaluation is performed over 3D volumes. 

\noindent \textbf{\textit{Evaluation.}} \textbf{Segmentation:} we employ common segmentation metrics in the medical domain, such as the DICE coefficient (DSC) and the 95$\%$ Hausdorff Distance (HD). \textbf{Calibration:} following recent works \cite{murugesan2023trust,murugesan2024neighbor} we resort to the expected calibration error (ECE) \cite{naeini2015obtaining} on foreground classes, as in \cite{islam2021spatially}, and Thresholded Adaptive Calibration Error (TACE) (threshold of $10^{-3}$) \cite{nixon2019measuring}. We further compute the Friedman rank \cite{friedman1937use}, to fairly compare the performance of different algorithms in various settings. More details are given in the \appen.

\vspace{-4mm}

\begin{table}[h!]
    
    \centering
    \scriptsize
    \caption{\textbf{Quantitative performance.} Discriminative (DSC $\uparrow$, HD $\downarrow$) and calibration (ECE $\downarrow$, TACE $\downarrow$) metrics, 
    using UNet as segmentation backbone. The best method is highlighted in bold, whereas the second best is underlined.}
    \begin{tabular}{l | cccc | cccc | c | c}
                        \toprule
                        & \multicolumn{4}{c|}{ACDC} & 
                        \multicolumn{4}{c|}{FLARE} & \multicolumn{1}{c|}{Friedman} & \multicolumn{1}{c}{Final}\\
                        \midrule
                        & DSC & HD & ECE  & TACE & DSC & HD & ECE  & TACE & Rank & Rank \\
        \midrule
    FL \cite{lin2017focal} ($\gamma=3$) & 0.620 & 7.30 & 0.153 &  0.224 & 0.834 & 6.65 & 0.053 &  0.145 & 7.88 & 8 \\
    ECP \cite{pereyra2017regularizing} ($\lambda=0.1$)  & 0.782 & 4.44 & 0.130  &  0.151 & 0.860 & \underline{5.30} & 0.037  & 0.134 & 5.38 & 7 \\
    LS \cite{szegedy2016rethinking} ($\alpha=0.1$)  & 0.809 & 3.30 & 0.083 &  0.093 & 0.860 & 5.33 & 0.055 &  0.050 & 4.88 & 4 \\
    SVLS$_\text{ IPMI'21}$ \cite{islam2021spatially} 
    & 0.824 & 2.81 & 0.091 & 0.138  & 0.857 & 5.72 & 0.039 & 0.144 & 5.25 & 5 \\
    MbLS$_\text{ CVPR'22}$ \cite{liu2022devil} 
    & 0.827 & 2.99 & 0.103 &  0.081 & 0.836 & 5.75 & 0.046 & 0.041 & 5.25 & 5 \\
    NACL$_\text{ MICCAI'23}$ \cite{murugesan2023trust}
    & \underline{0.854} & 2.93 & 0.068 &  0.073  & \underline{0.868} & \textbf{5.12}  & 0.033 & \textbf{0.031} & 2.25 & 2 \\
    BWCR$_\text{ MICCAI'23}$ \cite{karani2023boundary} & 0.841 & \underline{2.69} & \textbf{0.051} &  0.075 & 0.848 & 5.39 & \textbf{0.029} & 0.059 & 3.13 & 3 \\
    \textbf{\ourmethod (Ours)} & \textbf{0.877} & \textbf{1.72} & \underline{0.057} &  \textbf{0.058} & \textbf{0.876} & 5.52 & \textbf{0.029} &  \underline{0.033} & 1.75 & 1 \\
    \bottomrule
    \end{tabular}
    \label{table:main}
    \end{table}

\noindent \textbf{\textit{Comparison to state-of-the-art calibration approaches.}} In Table \ref{table:main} and \ref{table:main-nnUnet}, we present the quantitative results of our approach compared to a list of relevant state-of-the-art calibration approaches, when using UNet and nnUNet as segmentation backbones, respectively. In terms of \textbf{segmentation performance}, our proposed \ourmethod brings very competitive performance, typically ranking as best, or second best approach, regardless of the segmentation backbone employed. 
Regarding \textbf{calibration}, the trend observed is similar, with \ourmethod providing well-calibrated models, either improving or at par with state-of-the-art for calibration. Furthermore, as it is common in evaluating many methods in multiple settings, we assess the \textbf{overall performance} with a multi-criteria analysis, the Friedman Rank. The results from this metric, which are reported at the right-most columns of both Tables \ref{table:main} and \ref{table:main-nnUnet}, show that \ourmethod ranks at the first position, outperforming existing methods when a trade-off between calibration and segmentation performance is considered. Furthermore, the first rank position is maintained even when employing a more powerful backbone, i.e., nnUNet, 
consistently delivering the better segmentation-calibration compromise.

\begin{table}[h!]

\centering
\scriptsize
\caption{\textbf{Quantitative performance.} Discriminative (DSC $\uparrow$, HD $\downarrow$) and calibration (ECE $\downarrow$, TACE $\downarrow$) 
using nnUNet \cite{nnUNet} as segmentation backbone. The best method is highlighted in bold, whereas the second best is underlined.} \label{unet_results}
\begin{tabular}{l | cccc | cccc | c | c}
                    \toprule
                    & \multicolumn{4}{c|}{ACDC} & 
                    \multicolumn{4}{c|}{FLARE} & \multicolumn{1}{c|}{Friedman} & \multicolumn{1}{c}{Final}\\
                    \midrule
                    & DSC & HD & ECE & TACE & DSC & HD & ECE & TACE & Rank & Rank\\
    \midrule
FL \cite{lin2017focal} ($\gamma=3$) & 0.874 & 1.60 & 0.134 & 0.136  & 0.893 & 3.93 & 0.039 & 0.061 & 6.00 & 6 \\
ECP \cite{pereyra2017regularizing} ($\lambda=0.1$) & 0.889 & \underline{1.44} & 0.067 & 0.112  & 0.873 & 5.85 & 0.046 & 0.131 & 6.00 & 6 \\
LS \cite{szegedy2016rethinking} ($\alpha=0.1$) & \textbf{0.891} & \textbf{1.35} & 0.067 & 0.066 & 0.891 & 3.61 & 0.062 & 0.047 & 4.00 & 4\\
SVLS$_\text{ IPMI'21}$ \cite{islam2021spatially} 
& 0.883 & 1.69 & 0.059 & 0.111 & 0.894 & 4.02 & 0.026 & 0.115 & 5.13 & 5\\
MbLS$_\text{ CVPR'22}$ \cite{liu2022devil} 
& 0.886 & 1.46 & 0.057 & 0.052 & 0.891 & 3.65 & 0.031 & 0.031 & 3.50 & 3 \\
NACL $_\text{ MICCAI'23}$ \cite{murugesan2023trust}
& 0.884 & 1.52 & \underline{0.056} & 0.059  & \textbf{0.896} & \underline{3.34} & \textbf{0.025} & \textbf{0.026} & 2.50 & 2\\
BWCR$_\text{ MICCAI'23}$ \cite{karani2023boundary} & 0.864 & 1.82 & 0.063 &  0.079 & 0.868 & 4.47 & 0.041 & 0.099 & 6.63 & 8 \\
\textbf{\ourmethod (Ours)} & \textbf{0.891} & 1.48 & \textbf{0.052} & \textbf{0.051} & \underline{0.895} & \textbf{3.24} & 0.029 & \underline{0.029} & 1.88 & 1 \\
\bottomrule
\end{tabular}
\label{table:main-nnUnet}
\end{table}

\vspace{2mm}

\begin{figure}[h!]
     \centering
     \begin{subfigure}[b]{0.49\linewidth}
         \centering
         \includegraphics[width=\linewidth,trim={0.2cm 0.2cm 0.2cm 0.2cm},clip]{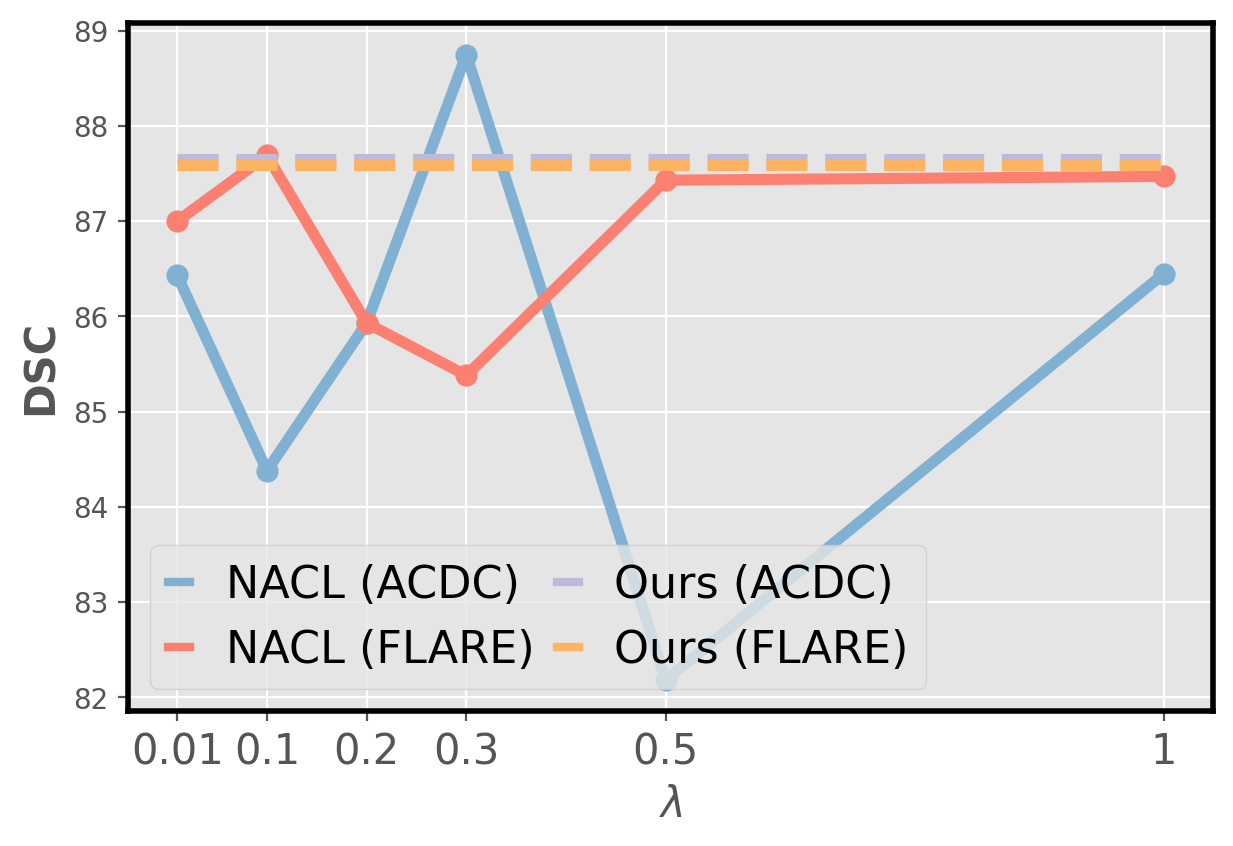}
     \end{subfigure}
     \begin{subfigure}[b]{0.49\linewidth}
         \centering
         \includegraphics[width=\linewidth,trim={0.2cm 0.2cm 0.2cm 0.2cm},clip]{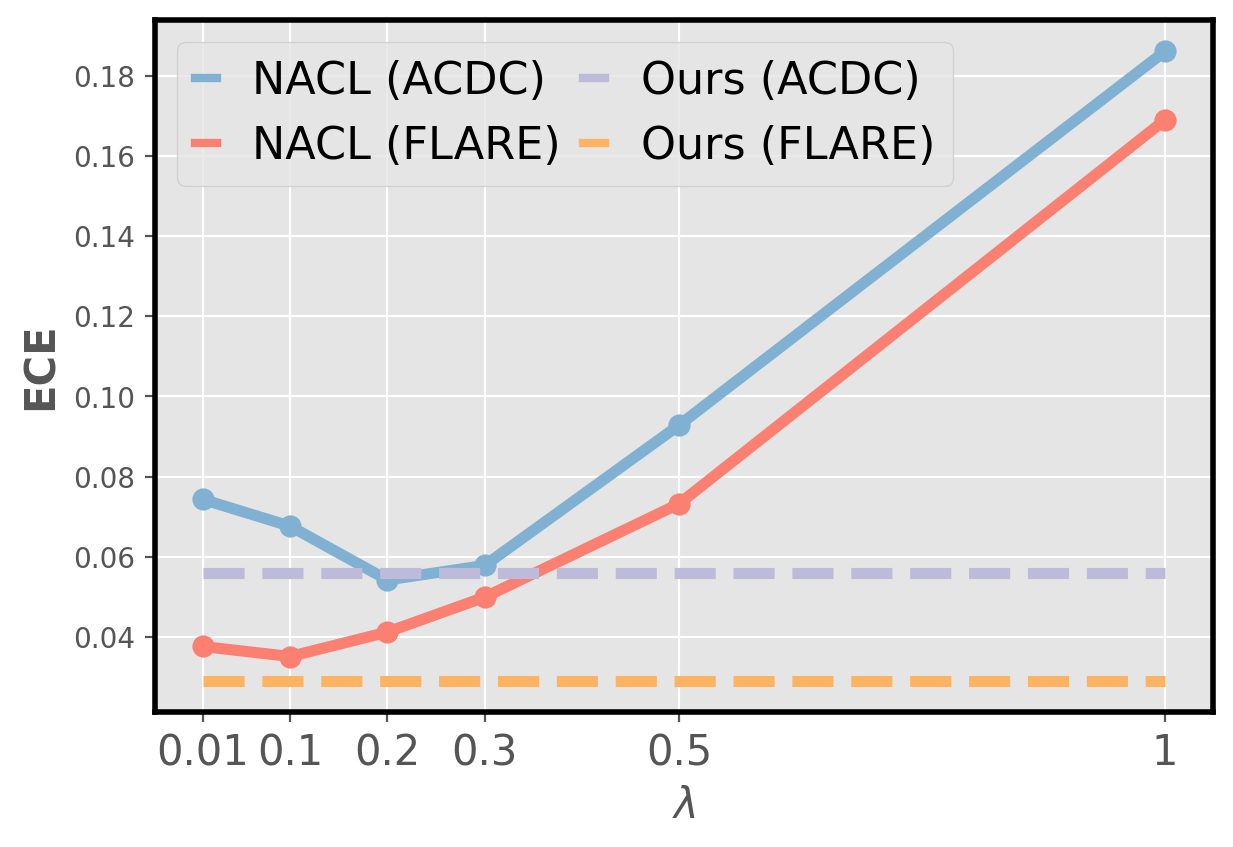}
     \end{subfigure}
    \caption{\textbf{Instability of NACL fine-tuning}. Discriminative (\textit{left}) \textit{vs.} calibration performance (\textit{right}) as a function of $\lambda$ in NACL \cite{murugesan2023trust}.}
    \label{fig:nacl-lambda}
\end{figure}

\vspace{-2mm}

\noindent \textbf{\textit{Benefits compared to NACL.}} In this section we compare the sensitivity of NACL \cite{murugesan2023trust} to the choice of its $\lambda$ value in Eq. $\ref{eq:constrained}$, as our approach improves NACL by incorporating a mechanism to learn and adapt the class and region-wise penalty terms $\lambda_{kr}$ in Eq. \ref{eq:proposed-ours}. We found that, despite performing at par in some settings, the performance of NACL significantly varies with the value of its penalty weight which, in addition, is dataset-dependent (Figure \ref{fig:nacl-lambda}). 
For example, the left plot demonstrates that while setting $\lambda=0.3$ in NACL yields the best discriminative performance in ACDC, it is substantially deteriorated in the FLARE dataset. Furthermore, the $\lambda$ value that optimizes the discriminative performance may not be the same that minimizes the miscalibration issue. Thus, while one may argue that by fine-tuning $\lambda$ in NACL can lead to improvements over \ourmethod (and only in certain settings), we advocate that performing a validation search in a dataset-basis is impractical for real-world problems, making of our approach an appealing solution.

\section{Conclusion.} We presented a novel approach to calibrate segmentation networks, which accounts for the inherent difficulties of different classes and regions. Results demonstrate that our approach outperforms existing approaches, becoming an excellent alternative to deliver high-performing and robust models.

\bibliographystyle{splncs04}
\bibliography{main}

\newpage 
\appendix
\setcounter{section}{0}

\section{Penalty functions for ALM: axioms}
\label{supp:subsection_penalties_axioms}

We provide here the requirements for a penalty function in the Augmented Lagrangian Multiplier (ALM) method.

A function $P : \real \times \real_{++} \times \real_{++} \rightarrow \real $ is a Penalty-Lagrangian function such that $P'(z, \rho, \lambda) \equiv \frac{\partial}{\partial z}P(z, \rho, \lambda)$ exists and is continuous for all $z \in \real$, $\rho \in \real_{++}$ and $\lambda \in \real_{++}$. In addition, a penalty function $P$ should satisfy the following four axioms \cite{birgin2005numerical}:
\begin{itemize}
    \item {\bf Axiom 1:} $P'(z, \rho, \lambda) \geq 0 \quad \forall z\in\real,  \rho \in \real_{++}, \lambda \in \real_{++}$
    
    \item {\bf Axiom 2:} $P'(0, \rho, \lambda) = \lambda  \quad \forall \rho \in \real_{++}, \lambda \in \real_{++}$
    
    \item {\bf Axiom 3:} If, for all $j\in\N, \; \lambda^{(j)} \in [\lambda_\text{min},\lambda_\text{max}]$, where $0 < \lambda_\text{min} \leq \lambda_\text{max} < \infty$, then:
    
    \qquad $\lim\limits_{j\rightarrow\infty}\rho^{(j)}=\infty$ and $\lim\limits_{j\rightarrow\infty}y^{(j)}>0$ imply that $\lim\limits_{j\rightarrow\infty}P'(y^{(j)}, \rho^{(j)}, \lambda^{(j)})=\infty$
    
    \item {\bf Axiom 4:} If, for all $j\in\N, \; \lambda^{(j)} \in [\lambda_\text{min},\lambda_\text{max}]$, where $0 < \lambda_\text{min} \leq \lambda_\text{max} < \infty$, then: 
    
    \qquad$\lim\limits_{j\rightarrow\infty}\rho^{(j)}=\infty$ and $\lim\limits_{j\rightarrow\infty}y^{(j)}<0$ imply that $\lim\limits_{j\rightarrow\infty}P'(y^{(j)}, \rho^{(j)}, \lambda^{(j)})=0$.
\end{itemize} 

\noindent While the first two axioms guarantee that the derivative of the Penalty-Lagrangian function $P$ \textit{w.r.t.} $z$ is positive and equals to $\lambda$ when $z=0$, the last two axioms guarantee that the derivative tends to infinity when the constraint is not satisfied, and zero otherwise. 

\section{Additional details on evaluation metrics}

\noindent \textbf{- Expectation Calibration Error (ECE).} 
ECE measures the correctness of the predictions by taking the weighted average of the error between accuracy and confidence. Let $N$ be the total number of data samples, $B$ be the total number of bins available for grouping. Then, ECE is given by: 
\begin{equation*}
\operatorname{ECE} = \sum_{b=1}^{B} \frac{n_b}{N} \left| \operatorname{acc}(b) - \operatorname{conf}(b) \right|,
\end{equation*}
Here, $n_{b}$, $acc(b)$, and $conf(b)$ denotes the number of samples, accuracy, and the confidence specific to that particular bin ($b$).  

\noindent \textbf{- Threshold Adaptive Calibration Error (TACE).} 
In order to get the best estimate of the overall calibration, it is better to focus on the bins where most of the prediction are made. This prevent the outcome to be skewed, and it is generally decided by an adaptive calibration range, give by $r$. Then, ACE is given by: 
\begin{equation*}
\operatorname{ACE} = \frac{1}{KR}\sum_{k=1}^K \sum_{r=1}^{R} \left| \operatorname{acc}(r, k) - \operatorname{conf}(r, k) \right|.
\end{equation*}
To further resolve the problem with tiny confidence scores making the calibration metric infinitesimal, TACE uses a threshold to skip the minimum ones.

\noindent \textbf{- Friedman rank.} This metric is employed when there exist multiple metrics and settings to compare several methods. It can be defined as $\operatorname{rank}_F=\frac{1}{m} \sum_{i=1}^m \operatorname{rank}_i$, with $m$ being the number of evaluation settings ($m=16$ in our work, 8 metrics $\times$ 2 datasets), and $\operatorname{rank}_i$ the rank of a method in the \textit{i}-th setting. Thus, the lower the rank obtained by an approach, the better this method is.

\section{Additional results: logit distributions}

\begin{figure}[h!]
     \centering
     \begin{subfigure}[b]{0.24\linewidth}
         \centering
         \includegraphics[width=\linewidth,trim={0.2cm 0.2cm 0.4cm 0.1cm},clip]{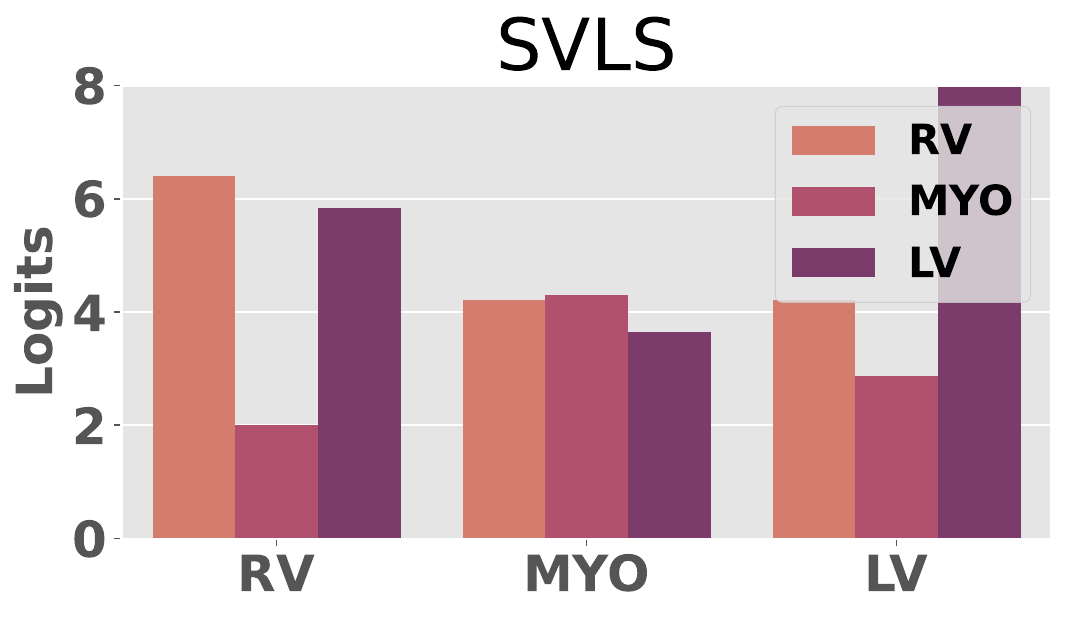}
     \end{subfigure}
     \begin{subfigure}[b]{0.24\linewidth}
         \centering
         \includegraphics[width=\linewidth,trim={0.2cm 0.2cm 0.4cm 0.1cm},clip]{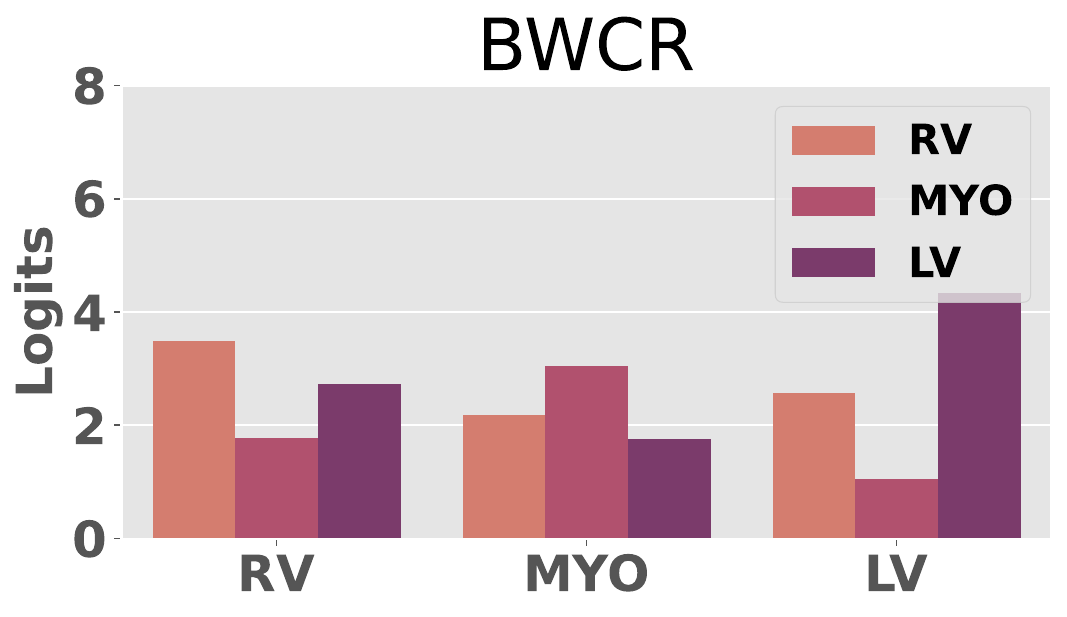}
     \end{subfigure}
     \begin{subfigure}[b]{0.24\linewidth}
         \centering
         \includegraphics[width=\linewidth,trim={0.2cm 0.2cm 0.4cm 0.1cm},clip]{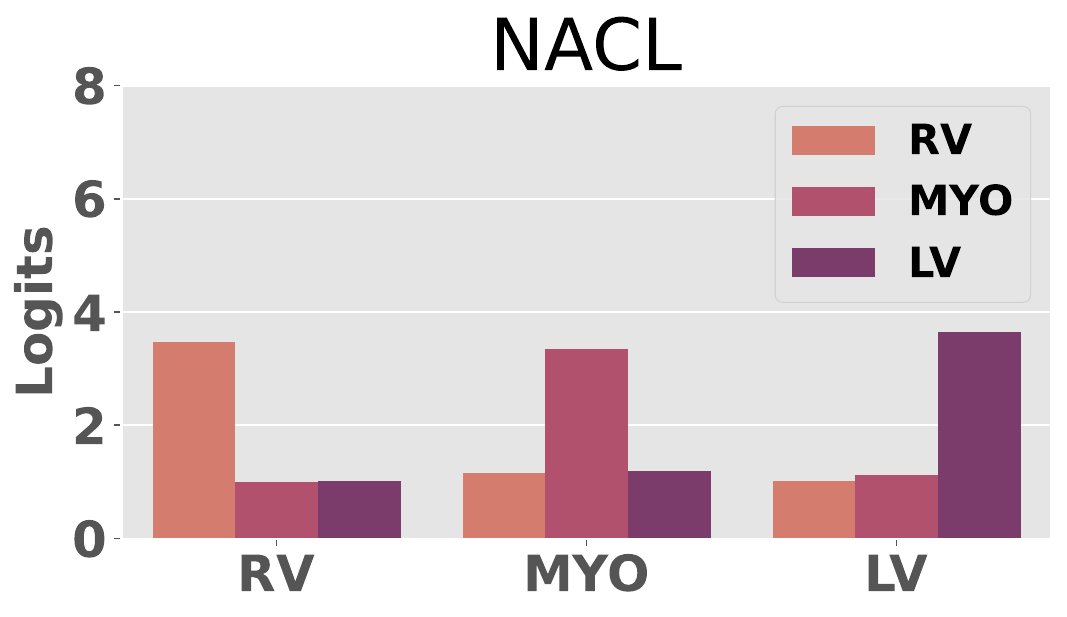}
     \end{subfigure}
     \begin{subfigure}[b]{0.24\linewidth}
         \centering
         \includegraphics[width=\linewidth,trim={0.2cm 0.2cm 0.4cm 0.1cm},clip]{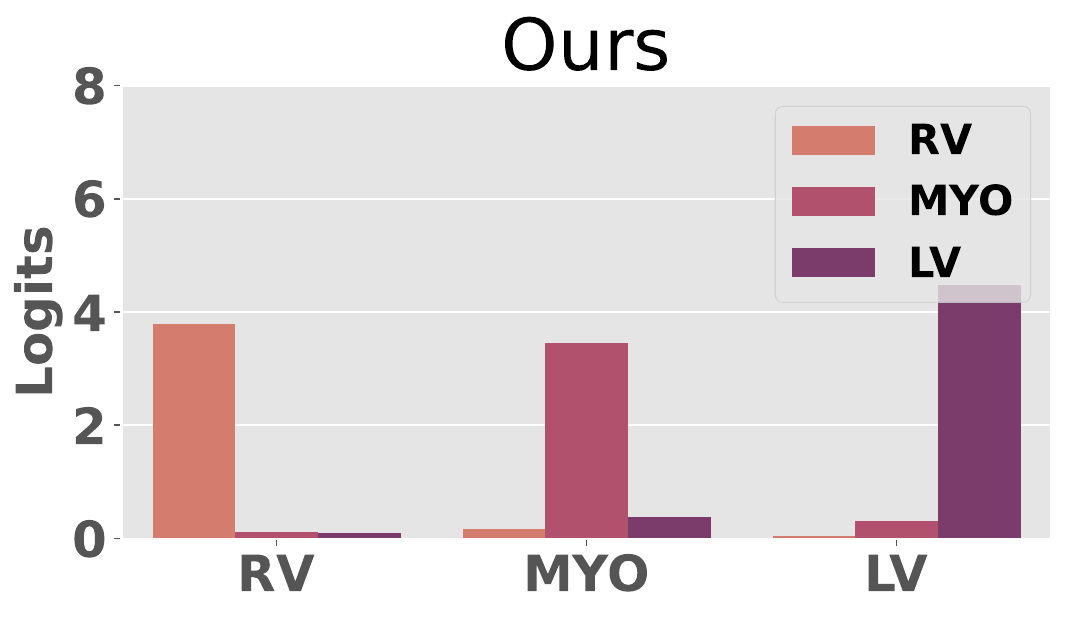}
     \end{subfigure}
     \begin{subfigure}[b]{0.24\linewidth}
         \centering
         \includegraphics[width=\linewidth,trim={0.2cm 0.2cm 0.4cm 0.1cm},clip]{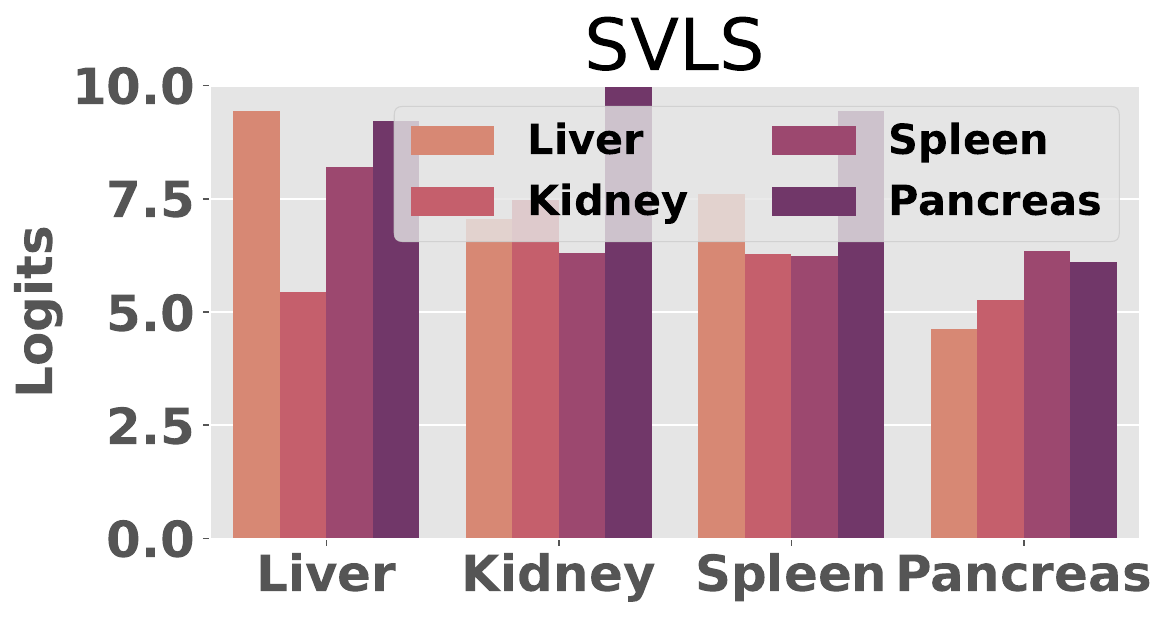}
     \end{subfigure}
     \begin{subfigure}[b]{0.24\linewidth}
         \centering
         \includegraphics[width=\linewidth,trim={0.2cm 0.2cm 0.4cm 0.1cm},clip]{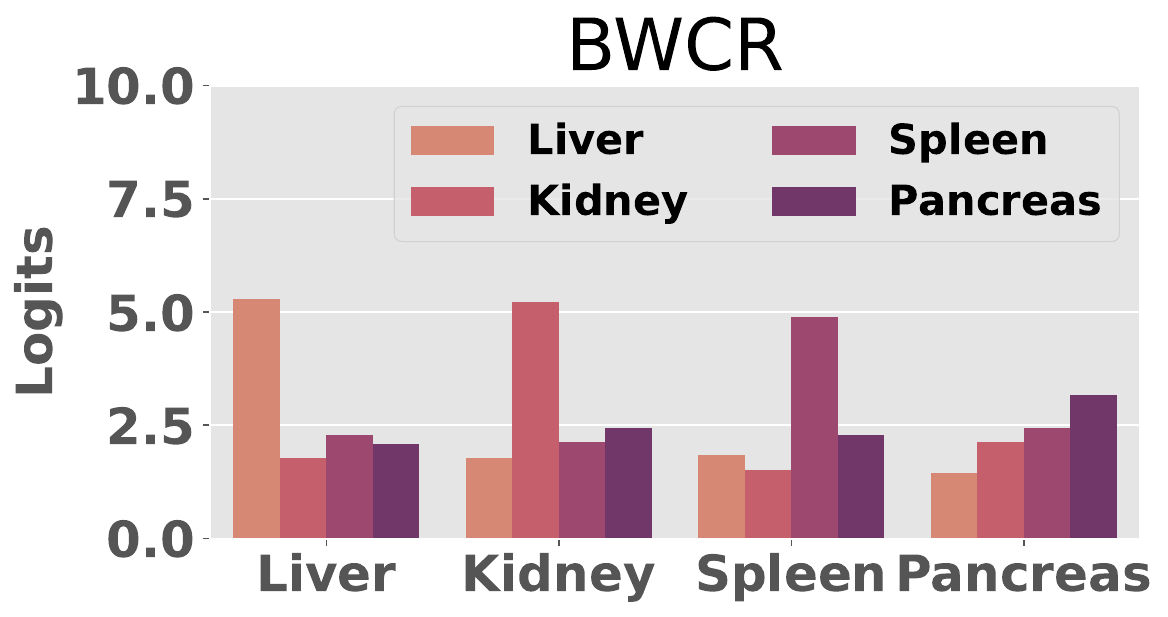}
     \end{subfigure}
     \begin{subfigure}[b]{0.24\linewidth}
         \centering
         \includegraphics[width=\linewidth,trim={0.2cm 0.2cm 0.4cm 0.1cm},clip]{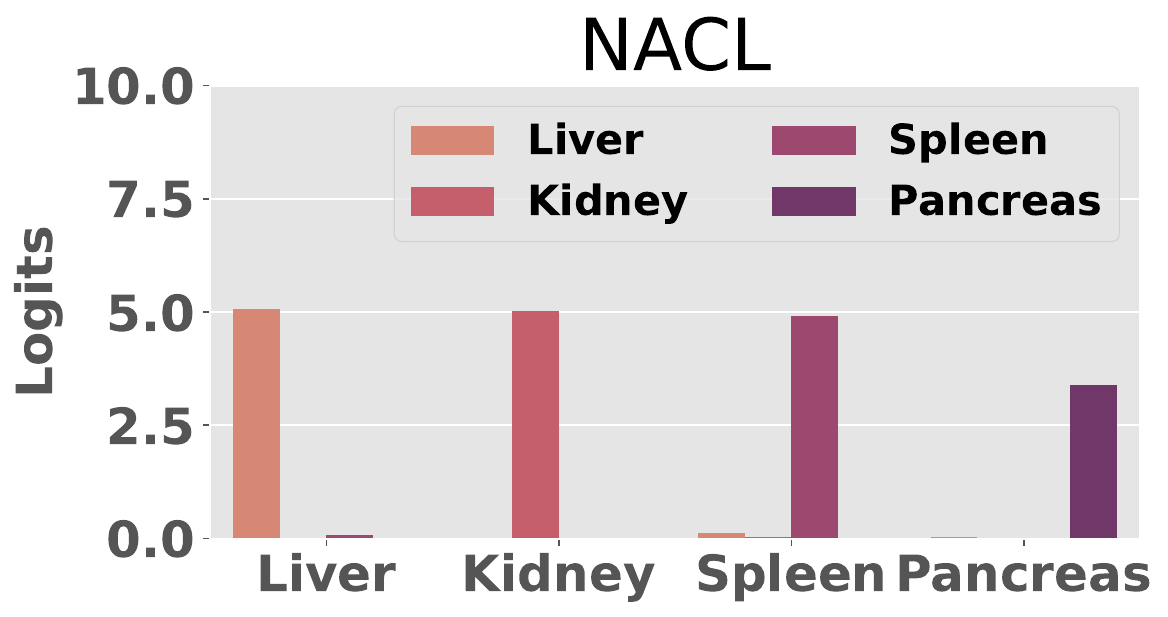}
     \end{subfigure}
     \begin{subfigure}[b]{0.24\linewidth}
         \centering
         \includegraphics[width=\linewidth,trim={0.2cm 0.2cm 0.4cm 0.1cm},clip]{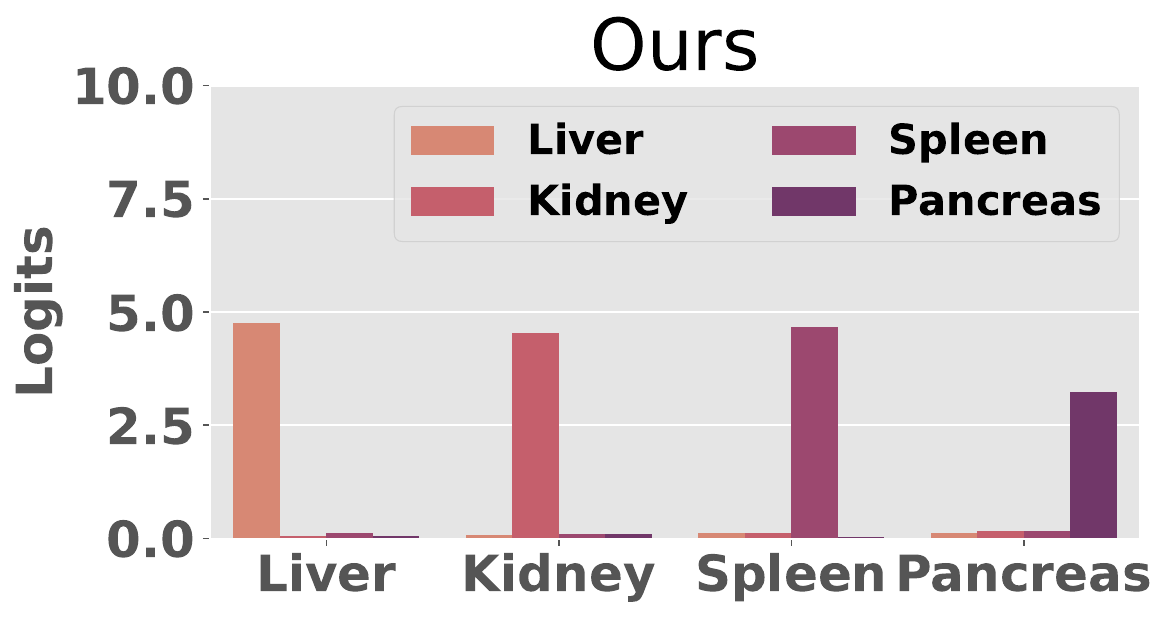}
     \end{subfigure}
    \caption{\textbf{Distribution of logit values.} 
    A desirable logit distribution exhibits lower winner logit magnitudes, which facilitate the training of a well-calibrated model, while pushing the remaining logit values to a considerable distance, and thus preserve a high discriminative power. An interesting observation from this figure is that, while NACL seems to generate desirable logit distributions for one dataset (FLARE), it may require fine-tuning of the $\lambda$ hyperparameter. In contrast, \ourmethod integrates an explicit mechanism to learn these values automatically, which facilitates a better compromise between segmentation and calibration.}
    \label{fig:logit-dist}
\end{figure}

\end{document}